\documentclass[letterpaper, 10 pt, conference]{ieeeconf}  

\IEEEoverridecommandlockouts                              

\usepackage{graphics} 
\usepackage{epsfig} 
\usepackage{amsmath} 
\usepackage{amssymb}  
\usepackage{multirow} 
\usepackage{booktabs} 
\usepackage{calc}
\usepackage{url}

\usepackage[ruled,linesnumbered]{algorithm2e}
\usepackage{algpseudocode}

\title{\LARGE \bf
TAPOM: Task-Space Topology-Guided Motion Planning for Manipulating Elongated Object in Cluttered Environments
}

\author{Zihao Li$^{1}$, Yiming Zhu$^{1}$, Zhe Zhong$^{1}$, Qinyuan Ren$^{1, \dag}$, Yijiang Huang$^{2}$
\thanks{$^{1}$Zihao Li, Yiming Zhu, Zhe Zhong, and Qinyuan Ren are with Collage of Control Science and Engineering, Zhejiang University, 310027, Hangzhou, China. Email: {\tt\small \{lzh\_jeong, flamezhu, smallcuz, renqinyuan\}@zju.edu.cn}.}%
\thanks{$^{2}$Yijiang Huang is with the Department of Computer Science, ETH Zurich, Switzerland. Email: {\tt\small yijiang.huang@inf.ethz.ch}.}%
\thanks{$^{\dag}$Corresponding author.}
}
\begin{document}

\maketitle
\thispagestyle{empty}
\pagestyle{empty}


\begin{abstract}
    Robotic manipulation in complex, constrained spaces is vital for widespread applications but challenging, particularly when navigating narrow passages with elongated objects. Existing planning methods often fail in these low-clearance scenarios due to the sampling difficulties or the local minima. This work proposes Topology-Aware Planning for Object Manipulation (TAPOM), which explicitly incorporates task-space topological analysis to enable efficient planning. TAPOM uses a high-level analysis to identify critical pathways and generate guiding keyframes, which are utilized in a low-level planner to find feasible configuration space trajectories. Experimental validation demonstrates significantly high success rates and improved efficiency over state-of-the-art methods on low-clearance manipulation tasks. This approach offers broad implications for enhancing manipulation capabilities of robots in complex real-world environments.

\end{abstract}


\section{Introduction}
\label{sec: introduction}

\begin{figure}[t]
    \centering
    \includegraphics[width=\columnwidth]{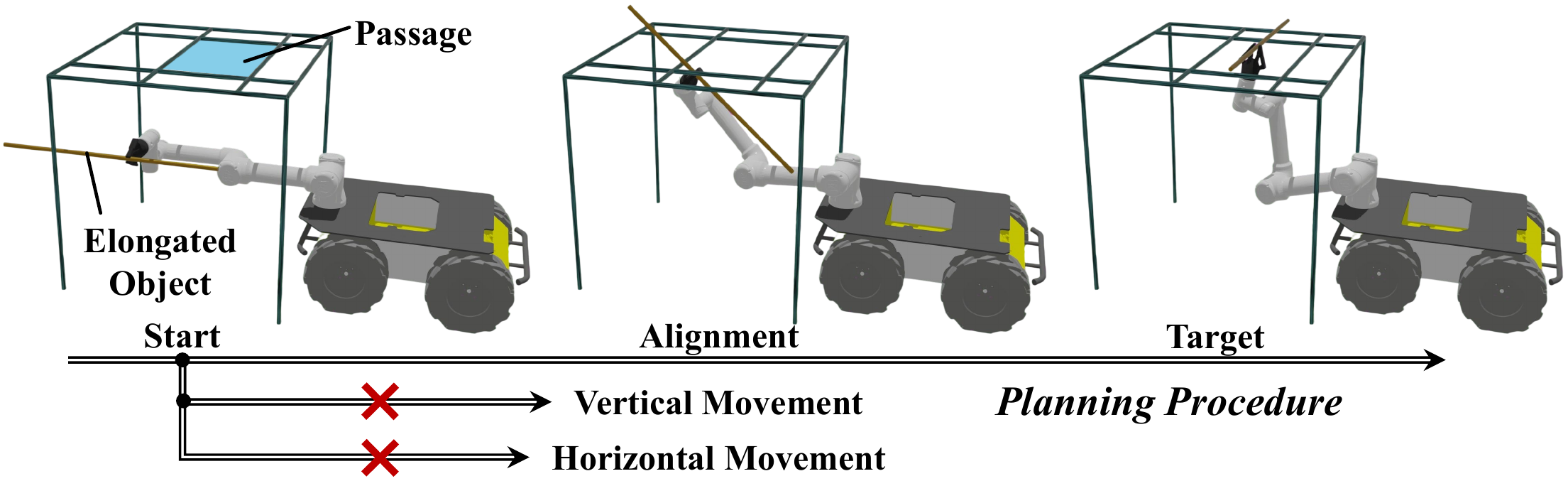}
    \caption{A typical manipulation task. A robot needs to manipulate an elongated object (a rebar beam) through a narrow passage (free sapces in a scaffold). Because translational motion is restricted, this robot needs to align object with the passage to pass through.}
    \label{fig: intro}
    \vspace{-2em}
\end{figure}

Growing demand for automation in manufacturing has driven widespread adoption of robotic manipulation. The complexity of motion planning increases significantly when robots are tasked with grasping and maneuvering objects with elongated geometries. Elongated objects are objects with a length that far exceeds their width or height, such as rods, beams, and pipes. Tasks, such as retrieving objects from cluttered shelves, assembling rebar into a rebar cage \cite{momeni2022automated}, and installing ceiling tiles \cite{liang2022trajectory}, often require robotic arms to operate more carefully than tasks without grasped objects. Figure \ref{fig: intro} illustrates a typical manipulation task: the robotic assembly of a rebar beam on a scaffold. Planning an elongated object through narrow openings is particularly challenging due to restricted translational mobility, necessitating complex rotations and out-of-plane maneuvers to achieve proper alignment. These additional motions expand the exploration space of the robots and introduce considerable difficulty for collision avoidance within the high-dimensional configuration space, namely C-space. These difficulties are characteristic of narrow passage problems in motion planning, where feasible paths are embedded in low-dimensional subsets of C-space and are difficult to discover efficiently.


Classical graph-based methods, e.g., Dijkstra's algorithm \cite{dijkstra_note_1959}, and sampling-based planners like PRM \cite{kavraki1996probabilistic} and RRT \cite{lavalle2001rapidly}, enable systematic or incremental exploration of C-space. More recent advancements include AIT* and EIT* \cite{strub2022adaptively}, which utilize asymmetric search and heuristic-informed estimates to improve exploration efficiency. However, their effectiveness is limited in narrow or low-clearance regions and requires trade-offs between resolution and computational cost. To alleviate these limitations, the optimization-based methods \cite{kalakrishnan2011stomp, schulman2013finding} refine trajectories without sampling, often producing smooth and efficient motions. Recently, researchers have concentrated on improving collision handling. Implicit SDF-based trajectory optimization \cite{wang2024implicit} eliminates the need for a fine joint resolution to prevent robots from colliding between discrete trajectory points. Nevertheless, these methods are prone to local minima, particularly in cluttered environments. To incorporate global structure and improve planning reliability, topology-based approaches, including methods leveraging homology classes \cite{bhattacharya2012topological}, constructing topological graphs \cite{yang2021graph}, and morphology-based approaches \cite{batista2023collision}, exploit the connectivity of C-space. But motion planning for manipulators grasping elongated objects remains difficult due to the complexity of identifying and connecting critical regions in constrained, high-dimensional spaces. Additionally, mapping task-space topology into configuration space introduces further non-trivial complexity.

In this paper, a hierarchical approach with task-space topology analysis and a keyframe-guided sampling-based planner, referred to as Topology-Aware Planning for Object Manipulation (TAPOM), is proposed. The core insight of TAPOM lies in decoupling the high-dimensional C-space search from the topology-aware analysis conducted in the low-dimensional task space. To deal with the identification of critical regions in cluttered and geometrically complex environments, task-space analysis is conducted by constructing a graph representing the connectivity of obstacles and extracting simple loops, named as ``channel'' in this paper. Key frames in C-space are generated by translating critical task-space channels into C-space. To tackle the challenge of low sampling probabilities within narrow passages, a keyframe-guided sampling-based planner is employed to generate feasible trajectories between keyframes. This low-level planner focuses on local connectivity of keyframes by growing and merging trees within relevant regions of C-space. This low-level planner is further equipped with a collision detection module to ensure the validity of generated trajectories, enhancing its robustness in producing collision-free motion plans within constrained spaces.

In summary, the key contributions of this work are as follows:

\begin{itemize}
    \item To explore topologically complex free spaces and identify critical pathways, task-space topology analysis is employed to explicitly model free space connectivity and find critical regions.
    \item Due to the sampling inefficiency encountered when planning through narrow passages in high-dimensional C-space, a keyframe-guided sampling-based planner is developed that leverages topological insights from high-level analysis to explore C-space.
    \item Experimental validation is conducted demonstrating the effectiveness and efficiency of proposed method compared to state-of-the-art planning baselines on manipulation tasks involving elongated objects and narrow passages.
\end{itemize}



Remainder of the article is organized as follows. Section \ref{sec: problem_formulation} formally defines the planning problem. Section \ref{sec: topology_aware_high_level_planning} details the proposed topology-aware high-level planning approach. In Section \ref{sec: low_level_planning}, the method for low-level path generation is presented. Section \ref{sec: experiments} describes experimental setup and results used to evaluate the performance of proposed method. Finally, Section \ref{sec: conclusion} provides a brief summary of the work and discusses directions for future research.

\begin{figure*}[h]
    \centering
    \includegraphics[width=\linewidth]{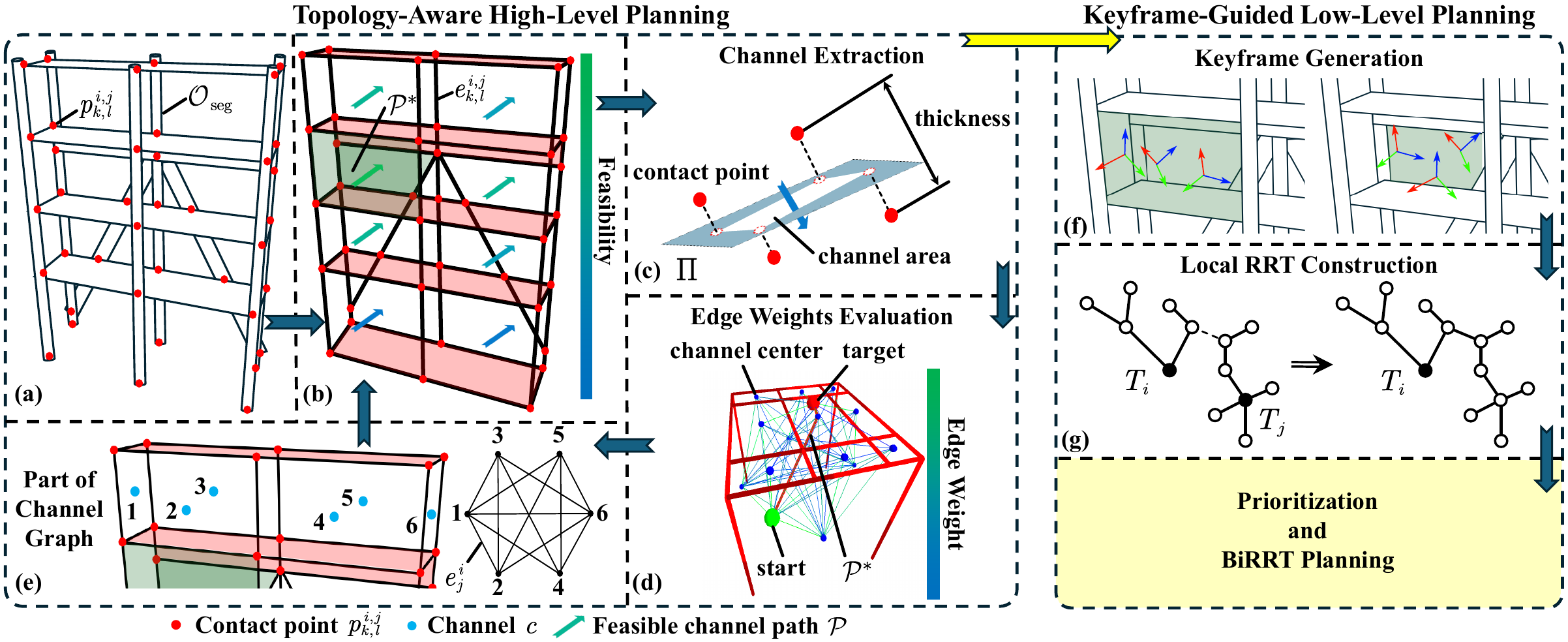}
    \caption{Overview of the proposed planner. \textbf{(a)} Obstacles are manually segmented into some sub-obstacles like boxes, cylinders, and spheres. Contact points between sub-obstacles are identified as red dots. \textbf{(b)} Connectivity graph $\mathcal{O}_{\text{seg}}$ is represented by nodes (red dots) and edges (black line). Simple loops (shaded areas, light red indicates invalid loops) are detected in this graph, representing potential channels. Arrows are several candidate channel paths with different feasibility. \textbf{(c)} Channel extraction: plane $\Pi$ is fitted to loop contact points $\{p_1, \dots, p_m\}$ via least-squares. Channel area (white) derived from convex hull on $\Pi$, thickness from perpendicular clearance. \textbf{(d)} Edge weights in channel graph $G_{\text{ch}}$: edge weights $w_{e_{ij}}$ indicate transition feasibility. Additionally, optimal high-level path $\mathcal{P}^*$ is highlighted in red. \textbf{(e)} Channel connectivity graph $G_{\text{ch}}$: nodes (blue dots) represent channels and edges represent feasible transitions between them. Channel paths in \textbf{(b)} are generated in this graph. \textbf{(f)} Generation of keyframes from the optimal high-level path $\mathcal{P}^*$ in task space. \textbf{(g)} Growing and merging trees within keyframe regions.}
    \label{fig: high_level_depict}
    \vspace{-1em}
\end{figure*}



\section{Problem Formulation}
\label{sec: problem_formulation}

Let the environment contain a set of static obstacles. Define
\begin{equation}
    \mathcal{O} = \{O_i\}_{i=1}^{n}
\end{equation}
as the set of all obstacles in the workspace. Each obstacle $O_i$ is decomposed into sub-obstacles by segmentation, so that
\begin{equation}
    \mathcal{O}_{\text{seg}, i} = \left\{o_{i, j} | o_{i, j} \subseteq O_i, \bigcup\limits_{j}o_{i,j}=O_i\right\}.
    \label{eqn: obstacle_segmentation}
\end{equation}
All sub-obstacles form
\begin{equation}
    \mathcal{O}_{\text{seg}} = \bigcup\limits_{i=1}^{n} \mathcal{O}_{\text{seg}, i}.
\end{equation}
The adjacency of sub-obstacles is represented by a graph $G_{\mathcal{O}_{\text{seg}}} = (V,E)$ with $V = \mathcal{O}_{\text{seg}}$ and
\begin{equation}
    E = \left\{(o_{i,j}, o_{k,l}) |\texttt{cnt}(o_{i,j} \cap o_{k,l})=1 \right\}.
\end{equation}
Simple cycles in $G_{\mathcal{O}_{\text{seg}}}$ correspond to potential passages. Each valid cycle defines a channel $c$ as the planar region bounded by the projected contact points of the cycle. Denote the set of all channels as
\begin{equation}
    \mathcal{C} = \{c_1, c_2, \dots, c_r\}.
\end{equation}

A channel connectivity graph $G_{\text{ch}} = (V_{\text{ch}}, E_{\text{ch}})$ is constructed where each $v_i \in V_{\text{ch}}$ represents channel $c_i \in \mathcal{C}$. An edge $e^{i}_{j} = (v_i, v_j) \in E_{\text{ch}}$ exists if channels $c_i$ and $c_j$ can be directly connected.

Define $\mathcal{Q} \subseteq \mathbb{R}^{d}$ as the robot's C-space and let $\mathcal{R}(q)$ be the robot geometry in configuration $q$. The collision-free subset is
\begin{equation}
    \mathcal{Q}_{\text{free}} = \{q \in \mathcal{Q} | \mathcal{R}(q) \cap \mathcal{O}_{\text{seg}} = \emptyset \}.
\end{equation}
Given a start configuration $q_{\text{start}}\in\mathcal{Q}_{\text{free}}$ and a goal configuration $q_{\text{goal}}\in\mathcal{Q}_{\text{free}}$, keyframes in $\mathcal{Q}_{\text{free}}$ are associated with channels. For each channel $c_i$, let $\mathcal{K}_{c_i} \subseteq \mathcal{Q}_{\text{free}}$ be its keyframe set, and let
\begin{equation}
    \mathcal{K} = \bigcup\limits_{c\in\mathcal{C}} \mathcal{K}_{c}
\end{equation}
be the set of all keyframes.

A high-level path $\mathcal{P}_{k} = (c_{k,1}, c_{k,2}, \dots, c_{k,m_k})$ is a connected sequence of channels in $G_{\text{ch}}$ from a channel near $q_{\text{start}}$ to one near $q_{\text{goal}}$. Each path $\mathcal{P}_{k}$ is assigned a score $W(\mathcal{P}_k)$. The optimal high-level path is
\begin{equation}
    \mathcal{P}^* = \arg \max\limits_{\mathcal{P}_k} W(\mathcal{P}_k).
\end{equation}
Once $\mathcal{P}^*$ and its keyframe sequence $\{q_k\}\subseteq\mathcal{K}$ are determined, the low-level planner can be formulated as
\begin{equation}
    \begin{aligned}
        \exists \;   & \mathcal{J}(\mathcal{P^*}): [0, 1] \to \mathcal{Q}                 \\
        \text{s.t. } & \mathcal{J}(t) \in \mathcal{Q}_{\text{free}}, \forall t \in [0, 1] \\
                     & \mathcal{J}(0) = q_{\text{start}}                                  \\
                     & \mathcal{J}(1) = q_{\text{goal}}                                   \\
                     & \mathcal{P}^* \subset \mathcal{J}.
    \end{aligned}
\end{equation}


\section{Topology-Aware High-Level Planning}
\label{sec: topology_aware_high_level_planning}


This section describes the proposed topology-aware high-level planning approach, which is detailed in Algorithm~\ref{alg:high_level_planning}.

\subsection{Obstacle Segmentation and Topological Graph Construction}

Obstacles are represented by $\mathcal{O}$. For topological analysis, each $O_i \in \mathcal{O}$ is manually decomposed into sub-obstacles $\mathcal{O}_{\text{seg},i}$ (Algorithm~\ref{alg:high_level_planning}, line 1). All sub-obstacles are merged into a single set $\mathcal{O}_{\text{seg}}$.

A pair $(o_{i,j}, o_{k,l}) \in \mathcal{O}_{\text{seg}}$ is defined as topologically connected if a unique geometric contact point $p^{i,j}_{k,l}$ exists between them. Segmentation should ensure that at most one contact point per pair is used for obstacle graph construction, and simple loop extraction is used later. The resulting connectivity graph $G_{\mathcal{O}_{\text{seg}}}$ (Algorithm~\ref{alg:high_level_planning}, line 2) has nodes $v_{i,j} \in V$ representing $o_{i,j} \in \mathcal{O}_{\text{seg}}$ and undirected edges $e^{i,j}_{k,l} \in E$ indicating connection via contact point $p^{i,j}_{k,l}$ (Fig. \ref{fig: high_level_depict} (b)). $G_{\mathcal{O}_{\text{seg}}}$ reflects environmental structure, in which simple loops indicate possibile channels that robots can pass through.

\begin{algorithm}[t]
    \caption{High-Level Planning}
    \label{alg:high_level_planning}
    \KwIn{Obstacles $\mathcal{O}$, start config $q_{\text{start}}$, goal config $q_{\text{goal}}$, max path length $L_{\text{max}}$, connect threshold $\epsilon$, parameters $\alpha, \beta, \gamma$}
    \KwOut{Optimal channel path $\mathcal{P}^*$}

    $\mathcal{O}_{\text{seg}} \leftarrow \texttt{Segment}(\mathcal{O})$\;
    $G_{\mathcal{O}_{\text{seg}}} = (V, E) \leftarrow \texttt{BuildTopoGraph}(\mathcal{O}_{\text{seg}})$\;

    $\mathcal{L} \leftarrow \texttt{DetectSimpleLoops}(G_{\mathcal{O}_{\text{seg}}})$\;
    $\mathcal{C} \leftarrow \texttt{ExtractChannels}(\mathcal{L}, \texttt{Filtering})$\;

    $G_{\text{ch}} = (V_{\text{ch}}, E_{\text{ch}}) \leftarrow (\emptyset, \emptyset)$\;
    \ForEach{$c_i \in \mathcal{C}$}{
        Compute $w_{\text{reach}}(c_i)$, $w_{\text{pass}}(c_i)$\;
        $v_i \leftarrow \texttt{CreateNode}(c_i, w_{v_i})$\;
        $V_{\text{ch}} \leftarrow V_{\text{ch}} \cup \{v_i\}$\;
    }

    $v_{\text{start}} \leftarrow \texttt{CreateNode}(q_{\text{start}})$\;
    $v_{\text{goal}} \leftarrow \texttt{CreateNode}(q_{\text{goal}})$\;
    $V_{\text{ch}} \leftarrow V_{\text{ch}} \cup \{v_{\text{start}}, v_{\text{goal}}\}$\;
    $\texttt{ConnectNodes}(v_{\text{start}}, v_{\text{goal}}, \mathcal{C}, G_{\text{ch}})$\;

    \ForEach{$\{c_i, c_j\} \subseteq V_{\text{ch}}$}{
        $\xi^{i}_{j} \leftarrow \texttt{SampleConnections}(c_i, c_j)$\;
        $d^{i}_{j} \leftarrow \texttt{Distance}(c_i, c_j)$\;
        Compute $w_{e^{i}_{j}}$ via Eq.~\eqref{eqn: edge_weight}\;
        \If{$w_{e^{i}_{j}} > \epsilon$}{
            $E_{\text{ch}} \leftarrow E_{\text{ch}} \cup \{e^{i}_{j}\}$\;
        }
    }

    $\{\mathcal{P}_k\} \leftarrow \texttt{BFS}(G_{\text{ch}}, v_{\text{start}}, v_{\text{goal}}, L_{\text{max}})$\;

    $W_{\text{max}} \leftarrow -\infty$\;
    $\mathcal{P}^* \leftarrow \emptyset$\;
    \ForEach{$\mathcal{P}_k \in \{\mathcal{P}_k\}$}{
        Compute score $W(\mathcal{P}_k)$ via Eq.~\eqref{eqn: path_scoring}\;
        \If{$W(\mathcal{P}_k) > W_{\text{max}}$}{
            $W_{\text{max}} \leftarrow W(\mathcal{P}_k)$\;
            $\mathcal{P}^* \leftarrow \mathcal{P}_k$\;
        }
    }
    \Return $\mathcal{P}^*$
\end{algorithm}

\subsection{Loop Detection and Channel Extraction}

To facilitate high-level motion planning, it is necessary to identify traversable regions. Loops serve as representations of such regions, as they capture closed sequences of obstacles that may support the passage of manipulators. However, geometric abstraction alone is insufficient. A more structured representation—termed a channel—is required to capture both spatial and relational properties of these regions. Channels differ from raw loops by embedding the contact configuration into a planar, polygonal form, providing a consistent interface for downstream reasoning.

Potential traversable regions are identified by extracting simple loops $\mathcal{L}$ from $G_{\mathcal{O}_{\text{seg}}}$ (Algorithm~\ref{alg:high_level_planning}, line 3). A simple loop is defined as a cycle without repeated nodes or edges, with each segment connected through distinct contact points. These loops are detected using Depth-First Search (DFS), prioritizing the smallest cycles originating from the start node. Not all identified loops correspond to feasible traversal options, as some may result from multiple contact points on a single obstacle—these are filtered out as multi-contact artifacts (e.g., red regions in Fig.~\ref{fig: high_level_depict} (b)). For each valid loop, associated contact points $\{p_1, \dots, p_m\}$ are extracted (Algorithm\ref{alg:high_level_planning}, line 4), and a supporting plane $\Pi$ is fitted. A channel $c_i$ is defined as the planar region bounded by the projection of these contact points onto $\Pi$, forming a polygon whose centroid serves as the channel center. The set of all valid channels is denoted by $\mathcal{C}$.

\begin{algorithm}[t]
    \caption{Low-Level Path Generation}
    \label{alg:low_level_path_generation}
    \KwIn{High-level plan $\mathcal{P}^*$, robot model $\mathcal{R}$, start config $q_{\text{start}}$, goal config $q_{\text{goal}}$, collision function $f(\cdot)$, keyframe number $N_{\text{key}}$, minimal budget $\mathcal{B}_{\text{min}}$, budget scale factor $\kappa$}
    \KwOut{Collision-free joint trajectory $\mathcal{J}$}

    $\mathcal{K} \gets \emptyset$ \; \label{alg:low_level_path_generation:keyframe_sampling_start}
    \ForEach{$c \in \mathcal{P}^*$}{
        $\mathcal{K}_c \gets \emptyset$ \;
        \While{ $|\mathcal{K}_c| < N_{\text{key}}$}{
            $p \gets \texttt{RandomSample}(c)$\;
            $q \gets \texttt{IK\_Solve}(p, \mathcal{R})$ \;
            \If{$q \ne null \; \&\& \; !f(p) \; \&\& \; !f(q)$}{
                $\mathcal{K}_c \leftarrow \mathcal{K}_c \cup \{q\}$ \;
            }
        }
        $\mathcal{K}_c \leftarrow \texttt{Select}(\mathcal{K}_c)$ \;
        $\mathcal{K} \leftarrow \mathcal{K} \cup \mathcal{K}_c$ \; \label{alg:low_level_path_generation:keyframe_sampling_end}
    }
    $\mathcal{K} \leftarrow \mathcal{K} \cup \{q_{\text{start}}, q_{\text{goal}}\}$ \; \label{alg:low_level_path_generation:add_start_goal}

    $\mathcal{T} \gets \emptyset$ \; \label{alg:low_level_path_generation:local_rrt_start}
    \ForEach{$c \in \mathcal{P}^*$}{
    $\mathcal{T}_c \gets \emptyset$ \;
    \ForEach{$q_k \in \mathcal{K}_c$}{
        $T_k \gets \texttt{RRT}(\mathcal{K}_c, q_k, c, \mathcal{R}, f(\cdot))$\;
            $\mathcal{T}_c \gets \mathcal{T}_c \cup \{T_k\}$\;
            }
            \ForEach{$\{T_{i}, T_{j}\} \subseteq \mathcal{T}_c$}{
                \If{$\texttt{Connect}(T_{i}, T_{j})$}{
                    $T_{i} \gets \texttt{Merge}(T_{i}, T_{j})$\;
                    $T_{j} \gets \texttt{Merge}(T_{j}, T_{i})$\;
                }
            }
        $\mathcal{T} \gets \mathcal{T} \cup \mathcal{T}_c$ \; \label{alg:low_level_path_generation:local_rrt_end}
    }
    $\mathcal{K} \gets \texttt{Prioritize}(\mathcal{K}, \mathcal{T})$\;

    $\mathcal{J} \gets \texttt{BiRRT\_Connection}(\mathcal{K}, \mathcal{R}, f(\cdot), \mathcal{B}_{\text{min}}, \kappa)$ \;

    \Return{$\mathcal{J}$} \;
\end{algorithm}

\subsection{Channel Graph Construction and High-Level Planning}

Channel connectivity graph $G_{\text{ch}}$ is constructed from $\mathcal{C}$ (Algorithm~\ref{alg:high_level_planning}, lines 5-22), where $v_i \in V_{\text{ch}}$ corresponds to $c_i \in \mathcal{C}$ (Fig. \ref{fig: high_level_depict} (e)) and edge $e^{i}_{j} \in E_{\text{ch}}$ represents connectivity, detailed later, between $c_i$ and $c_j$. $G_{\text{ch}}$ abstracts workspace connectivity for global planning and is built by assigning node and edge weights.

\subsubsection{Node Weight}
A composite weight $w_{v_i}$ is assigned to each node $v_i$ (Algorithm~\ref{alg:high_level_planning}, lines 6-10) to represent its suitability for robot traversal. This weight is a combination of reachability ($w_{\text{reach}}(c_i)$) and passability ($w_{\text{pass}}(c_i)$). Reachability is estimated from the collision-free contact likelihood between robots and $c_i$, determined via sampled configurations. Passability quantifies the geometric capacity of channels, scoring higher for larger areas, with a maximum score assigned to channels exceeding the size of the grasped object. The final node weight is computed by
\begin{equation}
    w_{v_i} = \alpha w_{\text{reach}}(c_i) + \beta w_{\text{pass}}(c_i),
    \label{eqn: node_weight}
\end{equation}
where $\alpha, \beta$ are tunable weights.

\subsubsection{Edge Weight and Graph Construction}
Valid edges $e^{i}_{j} \in E_{\text{ch}}$ between $v_i$ and $v_j$ are determined by spatial proximity and visibility as a proxy for collision-free transfer (Algorithm~\ref{alg:high_level_planning}, lines 15-22). Edge weight $w_{e^{i}_{j}}$ is formulated as
\begin{equation}
    w_{e^{i}_{j}} = \frac{\xi^{i}_{j}}{d^{i}_{j}},
    \label{eqn: edge_weight}
\end{equation}
where $\xi^{i}_{j}$ is the ratio of valid straight-line connections between sampled points in $c_i$ and $c_j$, and $d^{i}_{j}$ is the Euclidean distance between their centers. Edges are added to $G_{\text{ch}}$ only if $w_{e^{i}_{j}} > \epsilon$.

\begin{algorithm}[t]
    \caption{\texttt{BiRRT\_Connection}}
    \label{alg:biRRT_connection}
    \KwIn{Prioritized keyframe set $\mathcal{K}$, robot model $\mathcal{R}$, collision function $f(\cdot)$, minimal budget $\mathcal{B}_{\text{min}}$, budget scale factor $\kappa$}
    \KwOut{Collision-free joint trajectory $\mathcal{J}$ or $\emptyset$}

    $\mathcal{J} \gets \emptyset$ \; \label{alg:biRRT_connection:init_J}

    $\mathcal{S}_{\text{all}} \gets \texttt{GetAllPaths}(\mathcal{K})$ \; \label{alg:biRRT_connection:get_paths}

    \ForEach{$s \in \mathcal{S}_{\text{all}}$}{ \label{alg:biRRT_connection:path_iteration_start}
    $success \gets \text{True}$ \;
    $\mathcal{J}_s \gets \emptyset$ \;
    $\mathcal{B}_{\text{cur}} \gets \mathcal{B}_{\text{min}}$ \;

    \For{$i = 1$ \KwTo $|s|-1$}{ \label{alg:biRRT_connection:segment_iteration_start}
    $\mathcal{J}_{i,i+1} \gets \texttt{BiRRT}(s[i], s[i+1], \mathcal{R}, f(\cdot), \mathcal{B}_{\text{cur}})$ \;
    \If{$\mathcal{J}_{i,i+1} = \emptyset$}{ \label{alg:biRRT_connection:segment_fail}
        $success \gets \text{False}$ \;
        \textbf{break} \;
    }
    $\mathcal{J}_s \gets \mathcal{J}_s \cup \mathcal{J}_{i,i+1}$ \;
    $\mathcal{B}_{\text{cur}} \gets \kappa\mathcal{B}_{\text{cur}}$ \;
    }

    \If{$success$}{ \label{alg:biRRT_connection:path_success}
        $\mathcal{J} \gets \mathcal{J}_s$ \;
        \Return{$\mathcal{J}$} \;
    } \label{alg:biRRT_connection:path_iteration_end}
    }
    \Return{$\emptyset$} \;
\end{algorithm}

\subsubsection{Path Scoring and High-Level Planning}
Start ($q_{\text{start}}$) and goal ($q_{\text{goal}}$) configurations are incorporated by adding nodes $v_{\text{start}}$ and $v_{\text{goal}}$ to $G_{\text{ch}}$ and connecting them to relevant channels (Algorithm~\ref{alg:high_level_planning}, lines 11-14). Candidate paths $\{\mathcal{P}_k\}$ are obtained via Breadth-First Search (BFS) on $G_{\text{ch}}$ (Algorithm~\ref{alg:high_level_planning}, line 23), constrained by the maximal path length $L_{\text{max}}$. Each path $\mathcal{P} = (v_1, \dots, v_g)$ is evaluated using composite scoring function in Eq. \eqref{eqn: path_scoring} (Algorithm~\ref{alg:high_level_planning}, lines 26-32).
\begin{equation}
    W(\mathcal{P}) = \frac{\sum\limits_{i=1}^{g-1} w_{v_i} \cdot w_{e^{i+1}_{i}} + w_{v_g}}{|\mathcal{P}|^\gamma},
    \label{eqn: path_scoring}
\end{equation}
where $|\mathcal{P}|$ is the number of nodes, and $\gamma$ penalizes length. Path $\mathcal{P}^*$ with the maximum score is selected as the high-level plan (Algorithm~\ref{alg:high_level_planning}, lines 30), e.g., green channels in Fig.~\ref{fig: high_level_depict} (b) and red path in Fig.~\ref{fig: high_level_depict} (d), and passed to the low-level planner.


\section{Low-Level Path Generation Fusing Keyframe Sampling and RRT Variants}
\label{sec: low_level_planning}


This section details the low-level planning procedure, which is summarized in Algorithm~\ref{alg:low_level_path_generation} and Fig.~\ref{fig: high_level_depict} (f-g).

\subsection{Keyframe Sampling from High-Level Paths}

High-level plan $\mathcal{P}^*$ provides a sequence of channels. Representative configurations (keyframes) are sampled for each channel $c \in \mathcal{P}^*$ (Algorithm~\ref{alg:low_level_path_generation}, lines 2-13). This involves sampling cartesian pose $p$ within $c$ (Algorithm~\ref{alg:low_level_path_generation}, line 5) and converting them to joint configurations $q$ using an inverse kinematics solver ($\texttt{IK\_Solve}(\cdot, \mathcal{R})$, Algorithm~\ref{alg:low_level_path_generation}, line 6). A configuration $q$ is valid only if both its task-space pose $p$ and joint configuration $q$ are collision-free ($f(p)$, $f(q)$, Algorithm~\ref{alg:low_level_path_generation}, line 7). Valid $q$ are added to the keyframe set $\mathcal{K}_c$ (Algorithm~\ref{alg:low_level_path_generation}, line 8), aiming for $N_{\text{key}}$ such configurations. $\mathcal{K}_c$ is refined via $\texttt{Select}$ (Algorithm~\ref{alg:low_level_path_generation}, line 11), which is to select configurations that cover the largest area in C-space. The final keyframe set $\mathcal{K}$ aggregates keyframes from all channels and includes $\{q_{\text{start}}\}$ and $\{q_{\text{goal}}\}$ (Algorithm~\ref{alg:low_level_path_generation}, line 12).

\subsection{Local RRT Construction and Keyframe Prioritization}

Local connectivity within each channel $c \in \mathcal{P}^*$ is assessed (Algorithm~\ref{alg:low_level_path_generation}, lines 16-29). For each keyframe $q_k \in \mathcal{K}_c$, a RRT tree $T_k$ is constructed, rooted at $q_k$ and confined within keyframe-defined boundaries ($\texttt{RRT}(\cdot, \cdot, c, \mathcal{R}, f(\cdot))$, Algorithm~\ref{alg:low_level_path_generation}, line 19). Attempts are made to connect pairs of trees $(T_{i}, T_{j})$ within the same channel using $\texttt{Connect}$ (Algorithm~\ref{alg:low_level_path_generation}, line 23). Successful connections lead to tree merging via $\texttt{Merge}$ rooted at the root of the first tree (Algorithm~\ref{alg:low_level_path_generation}, lines 24-25). The resulting set of trees per channel is $\mathcal{T}_c$. Keyframes in $\mathcal{K}$ are then prioritized ($\texttt{Prioritize}(\mathcal{K}, \mathcal{T})$, line 30) based on their local exploration results (e.g., size of merged trees), guiding subsequent global connection.

\subsection{Global Keyframe Connection via Budgeted BiRRT}

The final stage connects prioritized keyframes across consecutive channels to form complete trajectory $\mathcal{J}$ from $q_{\text{start}}$ to $q_{\text{goal}}$ (Algorithm~\ref{alg:low_level_path_generation}, line 31, detailed in Algorithm~\ref{alg:biRRT_connection}). Keyframe paths $\mathcal{S}_{\text{all}}$ are generated by traversal and sorted by the accumulated score of each path (Algorithm~\ref{alg:biRRT_connection}, line 2). For each path $s$, consecutive keyframes $(s[i], s[i+1])$ are connected using $\texttt{BiRRT}(\cdot, \cdot, \mathcal{R}, f(\cdot), \mathcal{B}_{\text{cur}})$ (Algorithm~\ref{alg:biRRT_connection}, line 8) with an adaptively allocated computational budget $\mathcal{B}_{\text{cur}}$. The budget starts at $\mathcal{B}_{\text{min}}$ (Algorithm~\ref{alg:biRRT_connection}, line 6) and scales by $\kappa \ge 1$ upon successful connection of a segment ($\mathcal{B}_{\text{cur}} \leftarrow \kappa\mathcal{B}_{\text{cur}}$, Algorithm~\ref{alg:biRRT_connection}, line 14). If any segment connection fails, the path $s$ is discarded. The first path successfully connected end-to-end yields the trajectory $\mathcal{J}$ (Algorithm~\ref{alg:biRRT_connection}, lines 16-19). If no path is successful, an empty trajectory is returned (Algorithm~\ref{alg:biRRT_connection}, line 21).



\begin{figure*}[t]
    \centering
    \includegraphics[width=.9\linewidth]{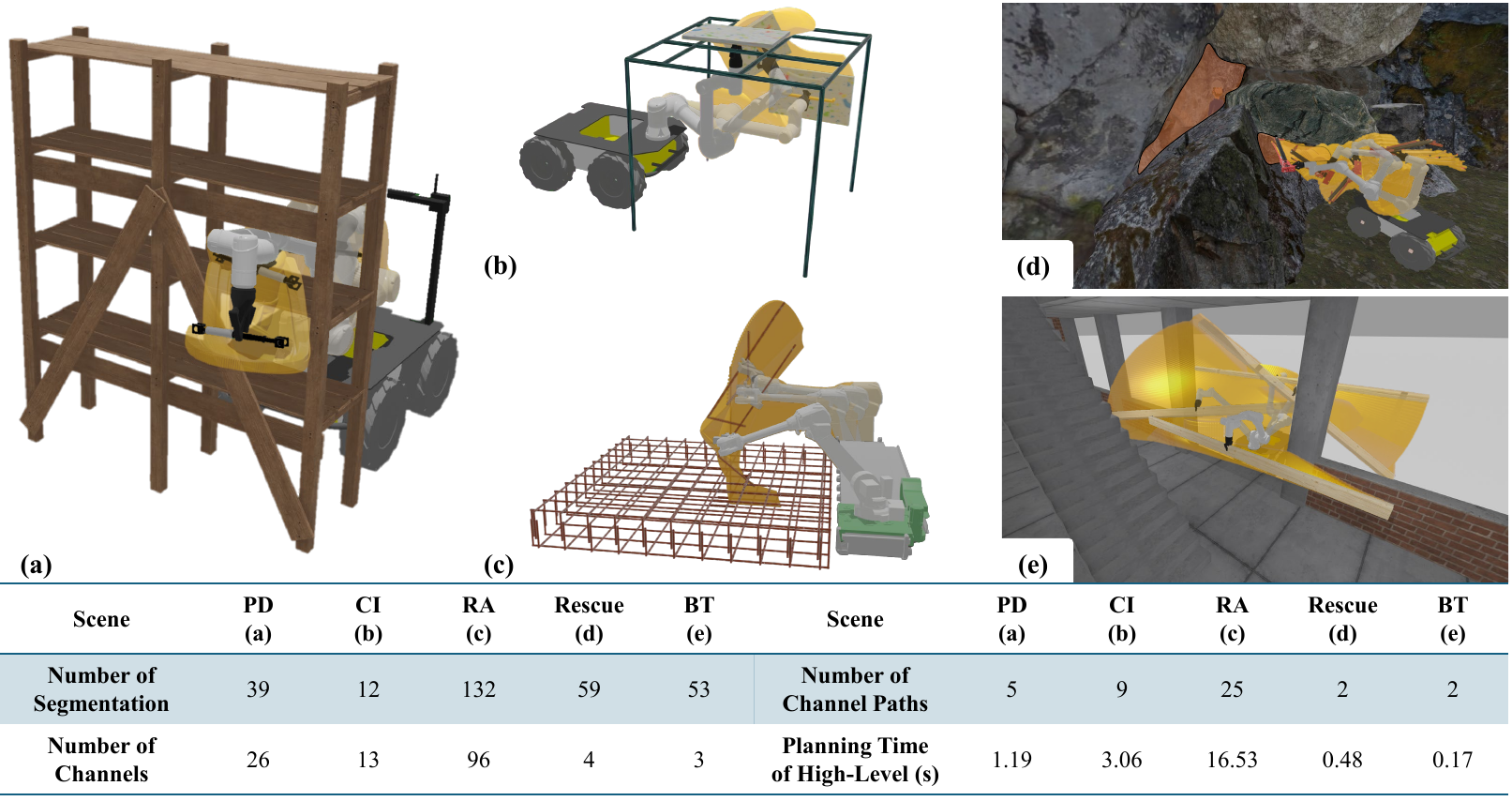}
    \caption{Performance of TAPOM in different scenarios. Light yellow area indicates the swept volume of robots and the grasped object, which starts from initial configuration and ends at goal configuration. \textbf{(a)} Part delivery through industrial shelving (\textbf{Part Delivery, PD}). \textbf{(b)} Installing a ceiling tile in a civil infrastructure (\textbf{Ceiling Installation, CI}), defined in \cite{liang2022trajectory}. \textbf{(c)} Robotic insertion of a rebar into a rebar cage typically used for rebar-reinforced concrete structures, inorder to weld a rebar to a skeleton (\textbf{Rebar Assembly, RA}), defined in \cite{momeni2022automated}. \textbf{(d)} Delivering a jack to a trapped man and helping him support potentially falling rocks (\textbf{Rescue}). Red areas indicate free space between rocks that enables robots to move. \textbf{(e)} Beam transportation in an unfinished building (\textbf{Beam Transportation, BT}).}
    \label{fig: results}
    \vspace{-1em}
\end{figure*}


\section{Simulations and Experiments}
\label{sec: experiments}


\subsection{Simulation Configurations}

All experiments are performed on a workstation equipped with a 12th Gen Intel Core i9-12900H CPU, an NVIDIA GeForce RTX 3060 GPU (utilized for GPU-accelerated planning methods), and 32 GB of RAM. Five distinct task scenarios are constructed to evaluate the planner in cluttered environments with narrow passages (see Fig.~\ref{fig: results}). Each scenario involves a 6-DOF robotic manipulator (either a UR5e or an ABB IRB 4600) performing a representative task: Beam Transportation (BT), Part Delivery (PD), Ceiling Installation (CI), Rescue, or Rebar Assembly (RA). These scenarios are specifically chosen to validate TAPOM's ability to navigate highly constrained spaces.

The proposed method is compared to several baseline motion planners to assess relative performance. The baseline set includes single-query planners (RRTConnect\cite{kuffner2000rrt}, EST\cite{hsu1997path}, STRIDE\cite{gipson2013resolution}, BFMT*\cite{starek2015asymptotically}), multi-query planners (PRM\cite{kavraki1996probabilistic}, LazyRRT\cite{rajasekaran2001handbook}), and optimizing planners (BIT*\cite{gammell2020batch}, EIT*\cite{strub2022adaptively}, and the GPU-accelerated cuRobo\cite{sundaralingam2023curobo}). Performance is evaluated using three metrics: planning success rate, average planning time, and average successful planning time. The planning success rate is defined as the proportion of trials that find a valid path within a $600.00s$ time limit. The average planning time is the mean computation time over all trials. The average successful planning time is the mean computation time considering only successful trials.

All baseline methods (except TAPOM and cuRobo) are run using the OMPL library \cite{sucan2012the-open-motion-planning-library} with parameters tuned for robust collision detection (state validity checking resolution $=5.0\times10^{-4}$, longest valid segment fraction $=5.0\times10^{-4}$, and planner range $=1.0\times10^{-2}$). CuRobo, the GPU-based planner, is executed with its default settings. The key parameters of TAPOM are empirically set to $\alpha=1.0$, $\beta=2.0$, $L_{\text{max}}=4$, $\gamma=3.0$, $N_{\text{key}}=20$, $\kappa=2$, $\epsilon=0.25$, and $\mathcal{B}_{\text{min}}=20s$. A uniform maximum planning time of $600.00s$ is enforced for all methods, and each scenario is executed with $10$ random trials to ensure consistency in evaluation\footnote{The code and data for the experiments are available at \url{https://github.com/Jeong-zju/TAPOM-Planner}}.

\subsection{Simulation Results}



Results in Fig.\ref{fig: main results} and trajectories in Fig.\ref{fig: results} are reported using three metrics. These metrics demonstrate TAPOM's effectiveness and efficiency compared to state-of-the-art baselines.

\subsubsection{Planning Success Rate}


Planning success rate serves as a key indicator of the adaptability to complex environments. A higher success rate reflects a stronger capacity to perceive structural and connectivity information in environments and to effectively utilize that information to generate valid paths. The middle figure of Fig.~\ref{fig: main results} reports success rates across all scenarios. TAPOM consistently achieves high success: $100\%$ in BT and Rescue, $90\%$ in PD and CI, and $80\%$ in RA. RRTConnect follows with $70\sim90\%$ across scenarios, while other baselines (e.g., EIT*, PRM, BIT*) either perform significantly worse or fail. The performance of TAPOM demonstrates not only its capability in challenging settings but also its effectiveness in capturing and leveraging topological features critical for successful motion planning.

\subsubsection{Average Planning Time}

\begin{figure*}[t]
    \centering
    \includegraphics[width=\linewidth]{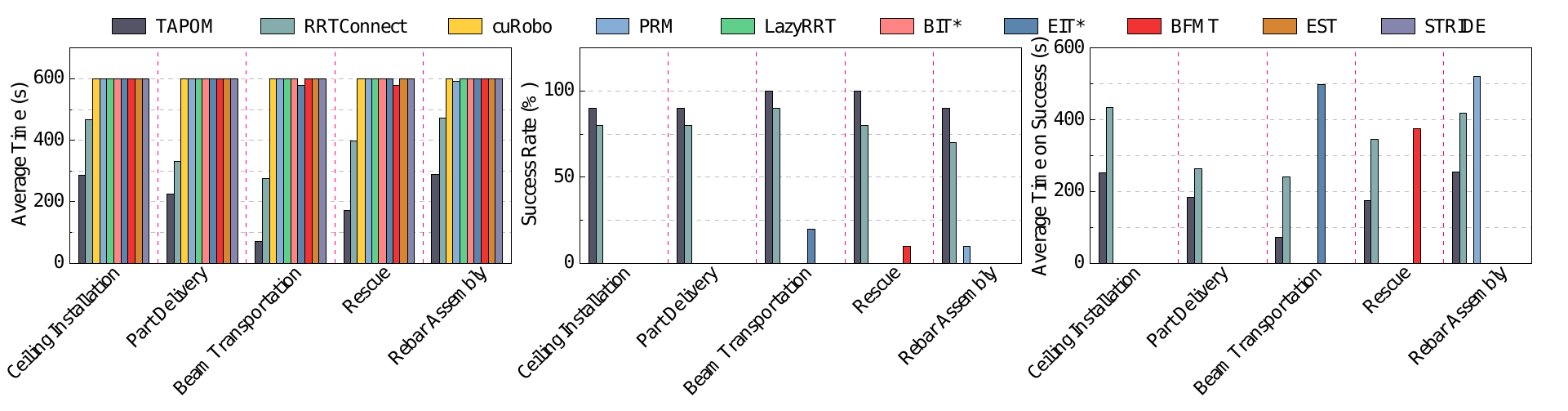}
    \caption{Results of experiments on different scenarios.}
    \label{fig: main results}
    \vspace{-1.5em}
\end{figure*}

Average planning time reflects the computational efficiency of a motion planning algorithm. As shown in the left plot of Fig.~\ref{fig: main results}, TAPOM requires substantially less time to find a solution compared to most baselines. Its average planning times are $71.62s$ (BT), $224.96s$ (PD), $287.07s$ (CI), $173.09s$ (Rescue), and $325.99s$ (RA). RRTConnect, the best-performing baseline, consistently requires $30\sim80\%$ more time, while other methods (e.g., BFMT*, PRM) take even longer or fail to complete planning.


For sampling-based approaches, time efficiency is closely tied to the characteristics of algorithms to explore C-space effectively. Faster planning times typically indicate more informed or structured exploration, particularly in high-dimensional or constrained settings. The consistently lower planning time of TAPOM across diverse scenarios suggests not only superior efficiency but also strong adaptability in both exploration and reasoning over structural constraints.



\subsubsection{Average Time on Successful Plans}

Specifically, the average time taken for successful planning reflects how quickly an algorithm can recognize key topological features, such as narrow passages or critical connections, and generate a feasible path accordingly. In unfamiliar or cluttered settings, a shorter success time often indicates stronger environment understanding, as planners can focus their exploration or optimization efforts on structurally significant regions. Although both average planning time and successful planning time relate to the capability of algorithms to adapt to complex environments, they emphasize different aspects. Average planning time captures a holistic view, whereas successful planning time highlights algorithms' efficiency in discovering and planning through constrained regions. For successful trials, TAPOM also maintains lower execution times, as shown in the right figure of Fig.~\ref{fig: main results}. It averages $71.62s$ (BT), $183.28s$ (PD), $252.30s$ (CI), $173.09s$ (Rescue), and $257.49s$ (RA). In comparison, RRTConnect ranges from $263.08s$ to $434.80s$, and PRM exceeds $500.00s$ in RA. The consistently lower success times of TAPOM demonstrate its capability to prioritize and exploit essential structural cues for motion generation.



\subsection{Experiment Results}

\begin{figure}[htbp]
    \centering
    \includegraphics[width=\columnwidth]{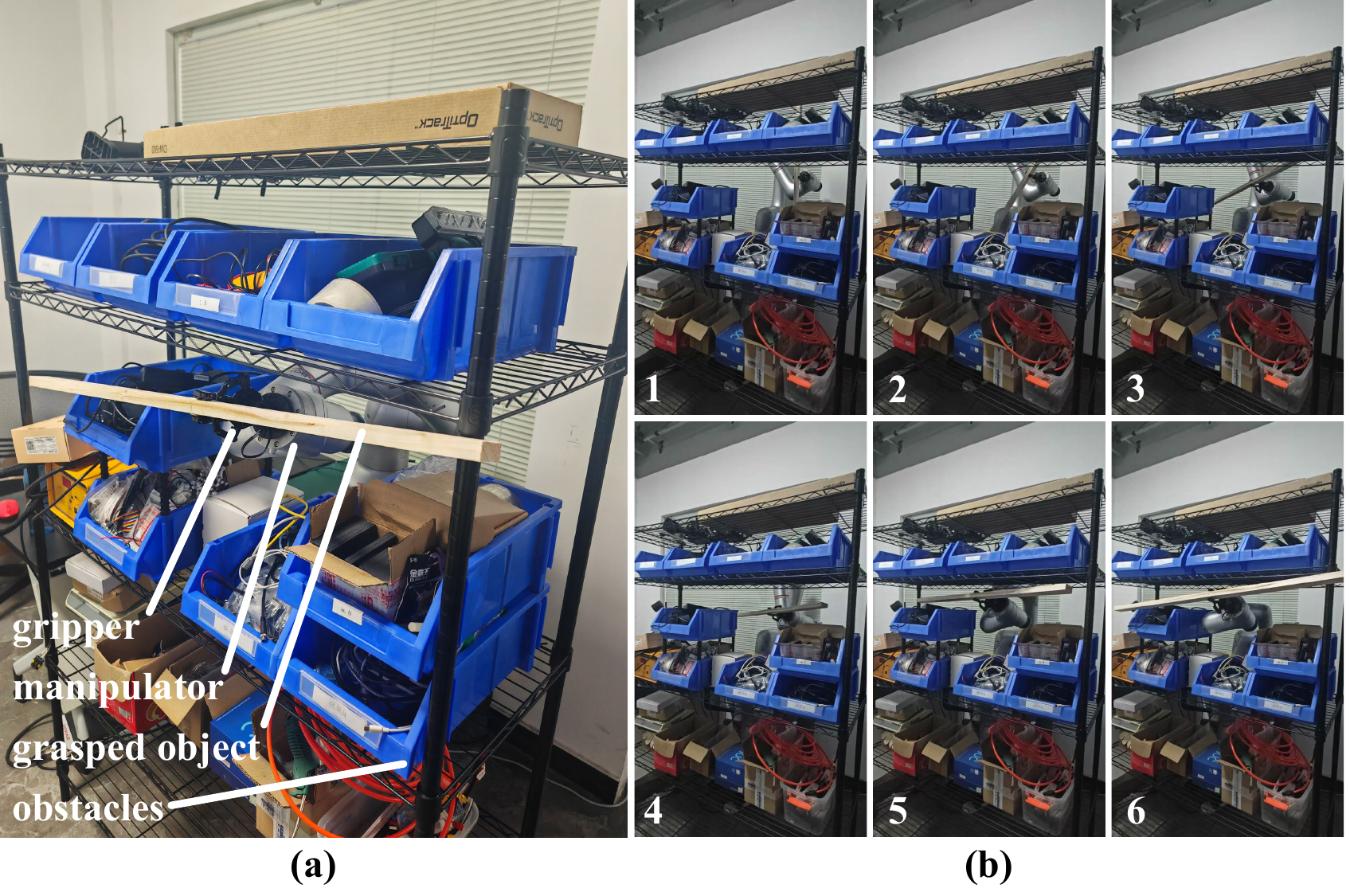}
    \caption{Results of real-world experiments. \textbf{(a)} The configurations of the real-world experiment. \textbf{(b)} The process of manipulation passing through a shelf full of devices and daily tools.}
    \label{fig: real}
    \vspace{-1em}
\end{figure}

Real-world experiments are conducted using a ROKAE CR7 manipulator to validate the planner in a cluttered environment. The setup consists of a shelf populated with common tools, providing only a narrow free space for maneuvering. An elongated wooden beam is grasped and transported through the shelf's constrained opening, where collisions are likely if any misalignment occurs due to limited clearance. A collision-free trajectory is obtained by the proposed method, avoiding contact with the shelf or any objects on it, which is shown in Fig.~\ref{fig: real}.

The differences between the real-world and simulation experiments are primarily attributed to hardware limitations. To ensure the robustness of the planner in real-world scenarios, conservative approaches are taken during the segmentation phase. Specifically, obstacles are segmented into smaller sub-obstacles, and their collision volumes are expanded to account for potential uncertainties in the environment and the robot's motion. In the real-world setup, disturbances in the rotation of the grasped object are observed due to factors such as structural instability, control inaccuracies, and external forces.

Table~\ref{tab: real} presents the results of the real-world experiments. The planning time is recorded as $186.29s$, with $92$ segmentations and $22$ channels identified in the environment. The rotation disturbance, $0.187$ radians, indicates the deviation in the object's orientation during the manipulation task. The translation length and rotation length of the trajectory are $0.637m$ and $1.793$ radians, respectively, demonstrating the total length of the motion.

\begin{table}[htbp]
    \centering
    \caption{Results of Real-World Experiments}
    \label{tab: real}
    \begin{tabular}{ccc}
        \toprule
        \textbf{Planning Time}        & \textbf{Num. of}            & \textbf{Num. of}         \\
        $(s)$                         & \textbf{Segmentation}       & \textbf{Channels}        \\ \midrule
        186.29                        & 92                          & 22                       \\ \toprule
        \textbf{Rotation Disturbance} & \textbf{Translation Length} & \textbf{Rotation Length} \\
        $(\text{rad})$                & $(m)$                       & $(\text{rad})$           \\ \midrule
        0.187                         & 0.637                       & 1.793                    \\
        \bottomrule
    \end{tabular}
\end{table}


As a result, the proposed planner successfully generates a collision-free trajectory that navigates the robot through the narrow passage, demonstrating its effectiveness in real-world applications. The consistency between the simulation and real-world results further validates the reliability of the planner in handling complex manipulation tasks in constrained environments.


\subsection{Ablation Studies}


\subsubsection{Module Ablation}

\begin{figure}[htbp]
    \centering
    \includegraphics[width=\columnwidth]{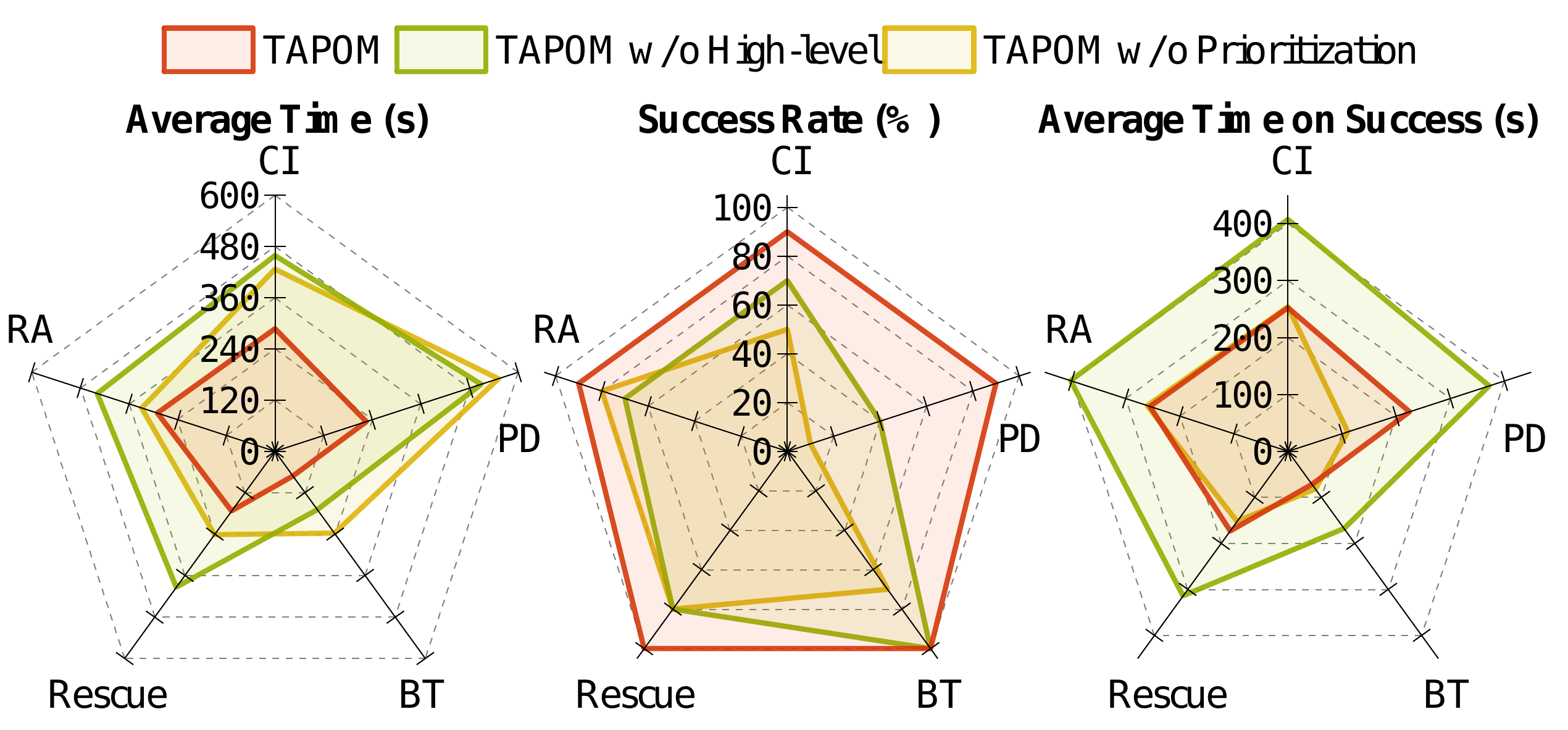}
    \caption{Results of module ablation.}
    \label{fig: ablation study module ablation}
\end{figure}

The removal of the High-Level Planning module reduces TAPOM to a BiRRT-like planner, resulting in lower success rates and increased planning times across all scenarios (see Fig.~\ref{fig: ablation study module ablation}). This highlights the crucial role of task-space topology analysis in explicitly modeling free space connectivity and identifying critical regions, which guides the search process more effectively in complex, cluttered environments. Ablating the Prioritization module impairs the low-level planner's ability to leverage topological insights for intelligent sub-task selection, diminishing exploration efficiency and significantly reducing success rates. This degradation demonstrates the importance of combining high-level topological reasoning with keyframe-guided sampling to mitigate the sampling inefficiency commonly encountered when navigating narrow passages in high-dimensional configuration spaces. Together, these results confirm that both modules are integral in realizing TAPOM's hierarchical framework.

\begin{figure}[htbp]
    \centering
    \includegraphics[width=\columnwidth]{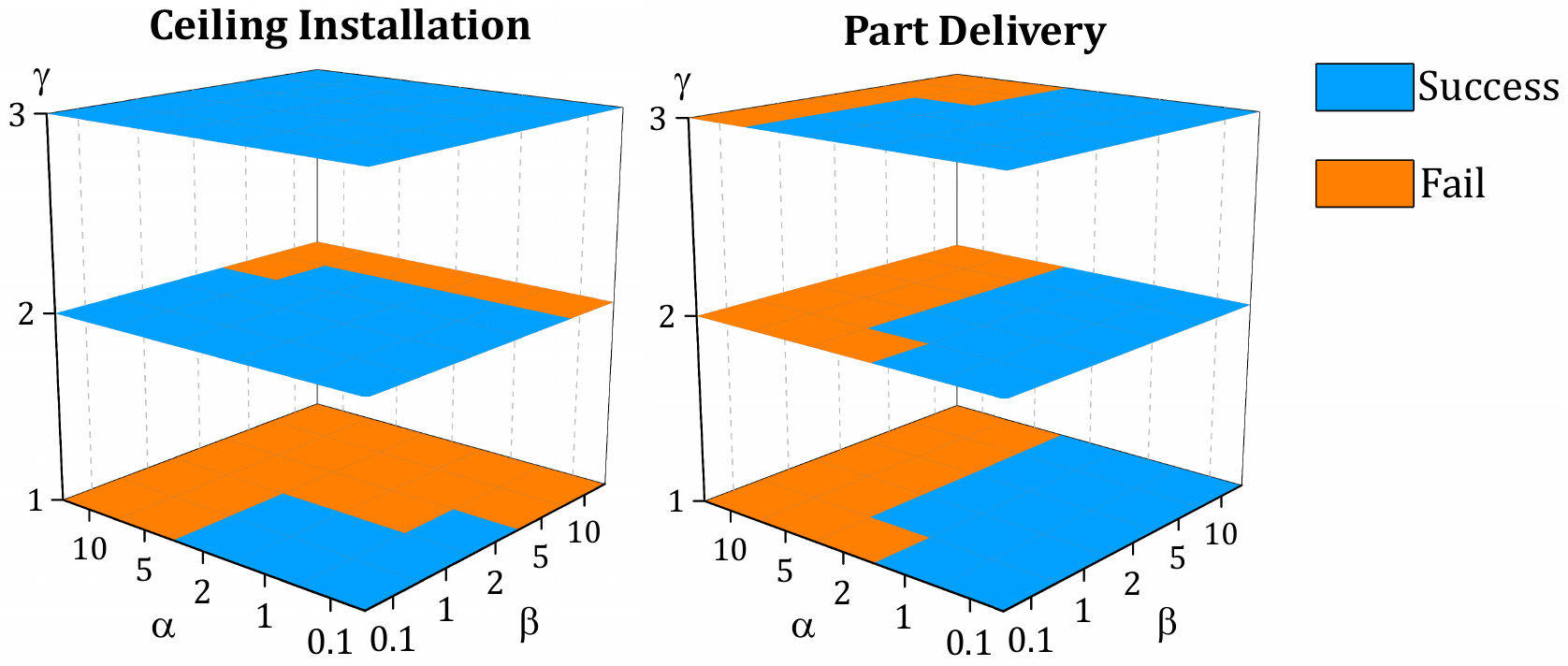}
    \caption{Results of parameter ablation.}
    \label{fig: ablation study param ablation}
\end{figure}

\subsubsection{Parameter Ablation}

Parameters $\alpha$, $\beta$, and $\gamma$ are evaluated to understand their impact on the effectiveness of the high-level topology-aware planning module (see Fig.~\ref{fig: ablation study param ablation}). Parameter $\gamma$, which penalizes path length, significantly influences path quality. An insufficient penalty (small $\gamma$) results in preference for excessively long paths that accumulate high scores within the channel graph, complicating subsequent low-level trajectory execution and decreasing overall planning success. Parameters $\alpha$ and $\beta$ represent reachability and passability weights, respectively, and exhibit a strong interactive effect on path selection. Imbalanced or excessively large values for these parameters distort the prioritization of transitions between critical regions, leading to suboptimal or invalid topological routes. Moreover, large values of $\alpha$ and $\beta$ reduce the relative influence of start and goal nodes in path scoring, which may hinder the identification of feasible keyframe sequences. Therefore, careful tuning of these parameters is essential to enable the high-level planner to accurately model free-space connectivity and guide the keyframe-guided sampling-based low-level planner.


\section{Conclusion}
\label{sec: conclusion}

A topology-aware hierarchical motion planner named TAPOM is proposed for complex manipulation in constrained environments. Topological decomposition and keyframe-guided trajectory planning are combined to exploit environmental structure and task semantics. Higher success rates and reduced planning times are demonstrated relative to state-of-the-art baselines in challenging scenarios. Limitations include lack of probabilistic completeness due to approximations in the high-level graph, dependence on clearly defined topological passages, and non-smooth or dynamically infeasible trajectories from RRT-based keyframe connections. Path optimization for smooth, dynamically feasible trajectories and perception-driven topology inference using vision-language models are planned to improve completeness and adaptability. The combination of topological abstraction and local planning is a promising foundation for scalable, structure-aware robot motion planning.

\addtolength{\textheight}{-12cm}   

\bibliographystyle{IEEEtran}
\bibliography{IEEEabrv,ref}

\end{document}